\documentclass[sigconf]{acmart}
\AtBeginDocument{%
  }

\setcopyright{acmlicensed}
\copyrightyear{2026}
\acmYear{2026}
\acmDOI{}
\acmISBN{}
\acmConference[KDD '26 SciSoc Agents \& LLMs]{the 32nd ACM SIGKDD Conference on Knowledge Discovery and Data Mining: SciSoc Agents \& LLMs Workshop}
{August 09--13, 2026}
{Jeju, Korea}
\acmBooktitle{the 32nd ACM SIGKDD Conference on Knowledge Discovery and Data Mining: SciSoc Agents \& LLMs Workshop (KDD~'26 SciSoc Agents \& LLMs), August 9--13, 2026, Jeju, Korea}

\usepackage[most]{tcolorbox}

\newtcolorbox{promptbox}[2][]{
  enhanced,
  breakable,
  colback=gray!3,
  colframe=black!55,
  coltitle=black,
  fonttitle=\bfseries\footnotesize,
  fontupper=\footnotesize,
  title={#2},
  boxrule=0.4pt,
  arc=1pt,
  left=4pt,
  right=4pt,
  top=3pt,
  bottom=3pt,
  before skip=4pt,
  after skip=4pt,
  #1
}

\begin{document}

\newcommand{\name}{SciLens}
\newcommand{\metricgain}[1]{\textcolor{green!50!black}{\scriptsize{}(#1)}}

\title{\name: Multi-modal Scientific Claim Verification with Agentic Entailment and Grounding}

\author{Yueming Wang}
\authornote{Equal Contribution.}
\affiliation{%
  \institution{The Hong Kong University of Science and Technology}
  \country{Hong Kong SAR, China}
}
\email{ywangtr@connect.ust.hk}

\author{Tianshi Zheng}
\authornotemark[1]
\affiliation{%
  \institution{The Hong Kong University of Science and Technology}
  \country{Hong Kong SAR, China}
}
\email{tzhengad@connect.ust.hk}

\author{Jiaxin Bai}
\affiliation{%
  \institution{Hong Kong Baptist University}
  \country{Hong Kong SAR, China}
}
\email{baijiaxin@hkbu.edu.hk}

\author{Yangqiu Song}
\affiliation{%
  \institution{The Hong Kong University of Science and Technology}
  \country{Hong Kong SAR, China}
}
\email{yqsong@cse.ust.hk}

\author{Ginny Wong}
\affiliation{%
  \institution{NVIDIA AI Technology Center}
  \country{Santa Clara, USA}
}
\email{gwong@nvidia.com}

\author{Simon See}
\affiliation{%
  \institution{NVIDIA AI Technology Center}
  \country{Santa Clara, USA}
}
\email{ssee@nvidia.com}

\renewcommand{\shortauthors}{Wang et al.}

\begin{abstract}
Scientific discovery increasingly relies on automated systems that generate
hypotheses, inspect multimodal evidence, and validate claims at scale. Yet
scientific claim verification is not well served by asking a vision-language
model for a direct binary judgment: claims often combine numerical results,
comparisons, scope qualifiers, and explanatory context, while evidence is
encoded in tables and figures with distinct grounding structures. We present
\textbf{\name}, an evidence-conditioned atomic entailment framework for
multimodal scientific claim verification. \name{} decomposes each claim into
central empirical atoms and background atoms, grounds the central atoms to
modality-specific evidence witnesses, and predicts the final label with an
atom-level entailment rule. For tables, atoms are grounded to rows, columns,
cells, arithmetic relations, and table scope; for figures, they are grounded
through panels, axes, legends, visual encodings, categories, trends, ranks, and
qualifier checks. This yields a unified validation procedure in which a claim is
supported only if every central empirical atom is entailed by the current
evidence. On the SciClaimEval development set, \name{} achieves 79.2\%
macro-F1 and 63.1\% pair accuracy, showing that structured agentic validation
improves both evidence sensitivity and interpretability.
\end{abstract}

\ccsdesc[500]{Computing methodologies~Natural language processing}
\ccsdesc[500]{Computing methodologies~Information extraction}
\ccsdesc[300]{Computing methodologies~Computer vision tasks}
\keywords{Scientific claim verification, multimodal reasoning, agentic workflow, evidence grounding, entailment}


\maketitle

\section{Introduction}

Agentic AI systems are increasingly used in scientific workflows, where
language models are combined with planning, tool use, retrieval, and
multimodal reasoning \cite{zheng2025automationautonomysurveylarge,luo2025llm4srsurveylargelanguage}. However, scientific progress depends not only on
generating hypotheses or fluent summaries, but also on verifying whether
reported claims are supported by the evidence in a paper. This problem is
challenging because evidence is often encoded in dense tables and
multi-panel figures, while claims combine exact values, comparisons,
scope qualifiers, and explanatory context.

SciClaimEval \cite{ho2026sciclaimevalcrossmodalclaimverification} provides a focused benchmark for this setting. A
system is given one scientific claim and one evidence item, either a table
or a figure, and must predict whether the evidence supports or refutes the
claim. Unlike generic visual question answering, this task requires
evidence-sensitive verification: a model must locate the relevant row,
column, panel, axis, legend entry, category, trend, rank, or statistical
cue, rather than relying on the caption or the general plausibility of the
claim.

We present \name, an agentic framework that formulates scientific table
and figure claim verification as a unified evidence-conditioned
entailment problem. \name{} decomposes each claim into checkable
empirical atoms, grounds those atoms to modality-specific evidence
witnesses, and predicts \textsc{Supported} only when the current table or
figure entails the central empirical content of the claim. The framework
is lightweight and task-aligned: it does not retrieve external documents,
compare against paired examples, or use mutation metadata. Instead, it
reasons only over the given claim, the current evidence item, and optional
local context provided with the sample.
\begin{figure*}[!t]
  \centering
  \includegraphics[width=\textwidth]{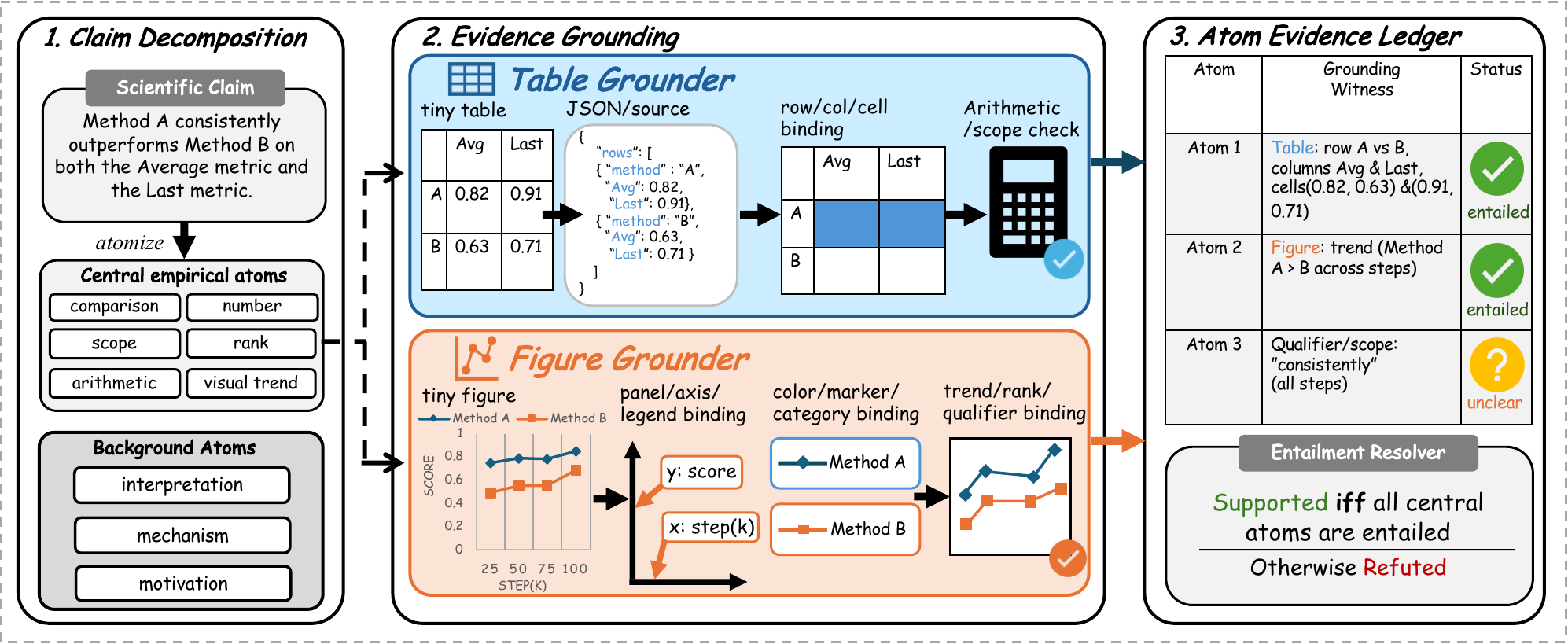}
  \caption{Overview of the \name{} framework.}
  \label{fig:scilens-framework}
\end{figure*}

On the SciClaimEval development set, \name{} achieves 79.2\% macro-F1
and 63.1\% pair accuracy, improving over the strongest vanilla
Qwen3-VL-30B-A3B baseline by 3.2 and 8.3 points, respectively. The gains
are especially large for pair accuracy, suggesting that atom-level
grounding improves evidence sensitivity rather than merely increasing
claim-level plausibility. These results show that structured agentic
validation can provide an effective and interpretable path toward
multimodal scientific claim verification.

\section{The SciLens Framework}
\label{sec:framework}

\subsection{Problem Formulation}

Given a claim \(c\), an evidence item \(e\), optional local context \(x\),
and modality \(m \in \{\mathrm{table}, \mathrm{figure}\}\), SciClaimEval
Subtask~1 requires a binary prediction
\(y \in \{\mathrm{Supported}, \mathrm{Refuted}\}\). We formulate the task
as evidence-conditioned entailment:
\[
\hat{y}=f_{\theta}(c,e,x,m).
\]
The prediction must be entailed by the current evidence item, rather than
by claim plausibility, caption-only cues, or evidence elsewhere in the
paper. \name{} operationalizes this principle through three steps:
decompose the claim into empirical atoms, ground each atom to
modality-specific evidence witnesses, and predict \textsc{Supported} only
when all central atoms are verified.

\subsection{Evidence Pack Construction}

For each instance, \name{} constructs an evidence pack containing the
claim, modality, caption, optional paragraph context, and evidence image.
For tables, the pack also includes structured table JSON and source-table
text when available, which help recover headers, cells, units, signs, and
OCR-sensitive content. These auxiliary representations are used only to
interpret the current table. For figures, the image remains the primary
evidence, while caption and context help identify panels, metrics, or
domain conventions without overriding the displayed visual content.

The pack is intentionally isolated: it excludes paired evidence items,
gold labels, mutation operations, and previous predictions. This prevents
pair-aware shortcuts and matches the evaluation protocol, where each instance is classified independently.

\subsection{Claim Decomposition}

The first stage decomposes the claim into central empirical atoms and
background atoms:
\[
D(c)=A_{\mathrm{central}}(c)\cup A_{\mathrm{bg}}(c).
\]
Central atoms are label-determining requirements, such as entities,
metrics, values, comparison directions, ranks, counts, arithmetic
relations, and statistical qualifiers. Background atoms
capture mechanisms, motivations, or explanatory statements that need not
be proven unless directly contradicted by the evidence. This separation
reflects scientific writing conventions: a table may establish that a
method ranks first without verifying the proposed explanation for why it
does so. The resolver therefore focuses on the empirical core while
avoiding over-penalization of explanatory phrasing.

\subsection{Table Grounding}

For table evidence, \name{} uses a three-stage workflow: claim
decomposition, evidence grounding, and entailment resolution. The table
grounder aligns each central atom to concrete witnesses, including row
labels, column headers, cells, units, signs, formatting cues, table
scope, and arithmetic. For claims involving ratios, averages,
differences, percentage points, fold changes, counts, or significance
statements, the grounder records the calculation or table cue used for
verification. We represent the witness for atom \(a_i\) as
\[
w_i^{\mathrm{tab}}=(r_i,h_i,v_i,u_i),
\]
where \(r_i\), \(h_i\), \(v_i\), and \(u_i\) denote the row, header, cell
value, and unit.

The table verifier treats exact numbers, named metrics, row--column
bindings, ranks, and hard scope terms such as ``all'', ``only'', and
``every'' as strict constraints. Approximate phrases such as ``around'',
``similar'', or ``slightly'' are accepted only when both direction and
magnitude are compatible with the table. This policy preserves ordinary
scientific approximations while enforcing truth-changing constraints.

\subsection{Figure Grounding}

Figure evidence requires visual rather than tabular witnesses. \name{}
therefore uses a reader--auditor--resolver workflow. The reader extracts
claim-critical visual requirements, including the relevant panel, metric,
axis direction, category relation, comparison, trend, rank, peak, numeric
relation, and qualifier. The auditor then binds entities and groups to
visual encodings in the current image, such as legends, colors, markers,
line styles, labels, panels, axes, and color bars, before assessing values
or trends. For figure atom \(a_i\), we denote the binding as
\[
B_i^{\mathrm{fig}}=\{(q,\ell_q):q\in Q_i\},
\]
where \(Q_i\) is the set of claim-critical visual slots and \(\ell_q\) is
the image element assigned to slot \(q\).

This binding step targets common failure modes in scientific figures:
swapped legends, incorrect panels, flipped axes, altered categories,
wrong trend directions, and missing significance markers. Visual mappings
mentioned in the claim or caption are treated as hypotheses to verify
against the image, not as facts. If a central visual binding is missing
or unreadable, \name{} predicts \textsc{Refuted} unless another explicit
visual cue verifies the relation.

\subsection{Typed Qualifier Policy}

\name{} applies a shared qualifier policy across modalities. Hard logical
qualifiers, such as ``all'', ``every'', ``only'', ``highest'', ``lowest'',
``best'', and ``same'', are checked strictly because a single
counterexample can falsify the claim. Soft descriptive qualifiers, such
as ``usually'', ``slightly'', ``similar'', and ``more pronounced'', are
judged by the dominant tabular or visual trend rather than by a fixed
threshold. Statistical qualifiers, including significance, \(p\)-values,
confidence intervals, and error bars, are central when explicitly invoked
by the claim and otherwise serve only as compatible supporting cues.

\subsection{Entailment Resolution and Guards}

The final resolver takes the evidence pack and same-instance stage notes
and outputs a JSON verdict with a short rationale. The decision rule is
conservative:
\[
\mathrm{Supported}(c,e) \iff
\forall a_i \in A_{\mathrm{central}}(c),\ e \models a_i .
\]
Thus, \name{} predicts \textsc{Refuted} if any central atom is
contradicted, missing, unclear, misaligned, or supported only by
caption/context. This atom-level rule yields an interpretable trace by
identifying the decisive evidence requirement rather than returning only
a binary label.

We add lightweight consistency guards to prevent later stages from
overriding grounded evidence. For tables, a \textsc{Refuted} prediction
is preserved when a central atom has an explicit contradiction; otherwise
unsupported refutations may be revised to \textsc{Supported}. For
figures, the resolver is invoked when earlier stages disagree or report
uncertainty, and explicit auditor contradictions are preserved against
generic supported flips. These guards are heuristic but use only
same-instance traces, avoiding leakage from labels, paired evidence, or
mutation metadata.
\begin{table*}[t]
  \centering
  \caption{Main results on the SciClaimEval development set. Macro-F1 is
  computed over individual claim--evidence samples, while pair accuracy
  is computed over paired evidence groups. All values are percentages;
  $n$ denotes the number of individual samples and $p$ denotes the
  number of paired evidence groups.}
  {\small
  \setlength{\tabcolsep}{5pt}
  \begin{tabular*}{\textwidth}{@{\extracolsep{\fill}}lcccccc@{}}
    \toprule
    Models
    & \multicolumn{2}{c}{Table}
    & \multicolumn{2}{c}{Figure}
    & \multicolumn{2}{c}{Overall} \\
    \cmidrule(lr){2-3} \cmidrule(lr){4-5} \cmidrule(l){6-7}
    & F1 & Pair Acc. & F1 & Pair Acc. & F1 & Pair Acc. \\
    & $n=482$ & $p=236$ & $n=265$ & $p=116$ & $n=747$ & $p=352$ \\
    \midrule
    \multicolumn{7}{l}{\textit{Vanilla LLM (baseline)}} \\
    \midrule
    InternVL3\_5-8B & 70.6 & 44.5 & 52.2 & 16.4 & 65.4 & 35.2 \\
    InternVL3\_5-14B & 72.0 & 49.2 & 59.7 & 25.9 & 68.5 & 41.5 \\
    InternVL3\_5-38B & 71.7 & 47.5 & 57.9 & 25.0 & 67.8 & 40.1 \\
    Qwen3-VL-4B & 74.2 & 49.2 & 63.7 & 32.8 & 70.6 & 43.8 \\
    Qwen3-VL-8B & 75.4 & 51.7 & 65.8 & 37.1 & 72.1 & 46.9 \\
    Qwen3-VL-30B-A3B & \underline{80.6} & \underline{62.7} & 66.6 & 38.8 & 76.0 & 54.8 \\
    \midrule
    \multicolumn{7}{l}{\textit{\name{} Agent}} \\
    \midrule
    Qwen3-VL-8B & 80.5 \metricgain{+5.1} & 61.9 \metricgain{+10.2} & \underline{67.7} \metricgain{+1.9} & \underline{42.2} \metricgain{+5.1} & \underline{76.2} \metricgain{+4.1} & \underline{55.4} \metricgain{+8.5} \\
    Qwen3-VL-30B-A3B & \textbf{81.7} \metricgain{+1.1} & \textbf{67.4} \metricgain{+4.7} & \textbf{74.3} \metricgain{+7.7} & \textbf{54.3} \metricgain{+15.5} & \textbf{79.2} \metricgain{+3.2} & \textbf{63.1} \metricgain{+8.3} \\
    \bottomrule
  \end{tabular*}}
  \label{tab:main-results}
\end{table*}
\begin{figure*}[t]
  \centering
  \includegraphics[width=\textwidth, keepaspectratio]{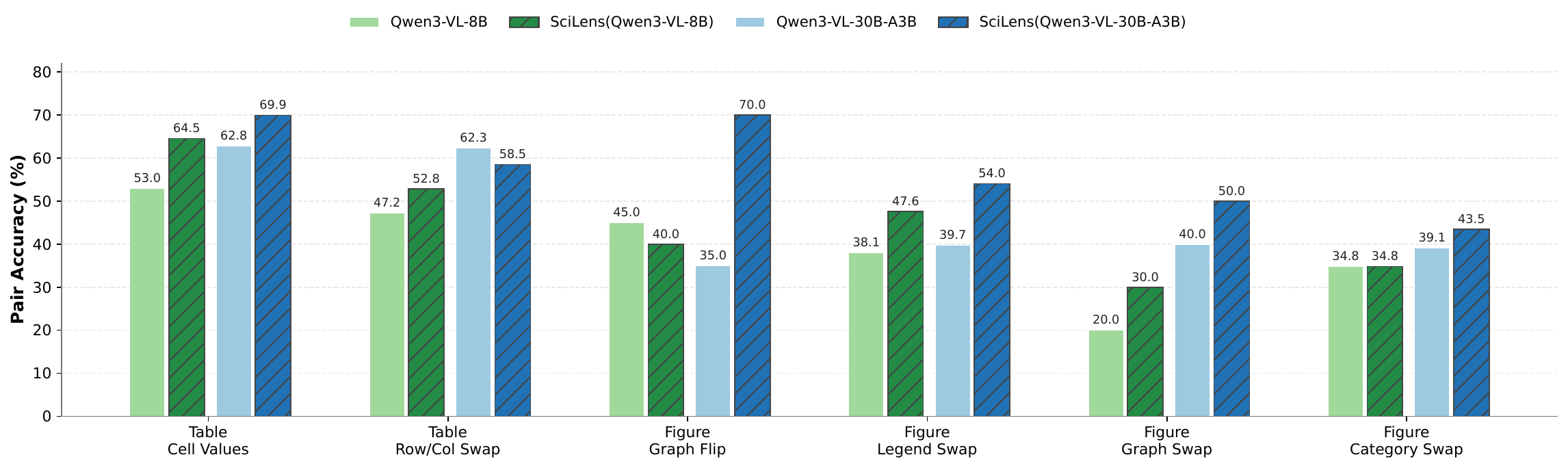}
  \caption{Operation-level comparison between \name{} agents and vanilla
  Qwen3-VL baselines.}
  \Description{Operation-level comparison plots showing SciLens agent
  performance against Qwen baselines across evidence modification types.}
  \label{fig:operation-level-comparison}
\end{figure*}
\section{Experiments}
\label{sec:experiments}

\subsection{Dataset}

We evaluate on the SciClaimEval Subtask~1 development set, which contains
747 claim--evidence instances: 482 table instances and 265 figure
instances. Each instance consists of a scientific claim, a table or figure
evidence image, a caption, optional local context, and a binary label
\(y \in \{\mathrm{Supported}, \mathrm{Refuted}\}\). The split contains
395 Supported and 352 Refuted instances.

SciClaimEval also groups many instances into paired examples through
\texttt{claim\_id\_pair}. Pair accuracy counts a group as correct only
if all predictions in that group are correct. Following the official
evaluation setting, instances marked as \texttt{no pair} are excluded
from this metric, yielding 236 table pairs and 116 figure pairs.

\subsection{Models and Settings}

We compare \name{} against vanilla multimodal LLM baselines. In the
vanilla setting, a single prompt asks the model to inspect the claim and
the current evidence item and output a JSON label. In the agentic
setting, the same prediction is produced by the staged framework in
\autoref{sec:framework}, using only same-instance evidence and
intermediate stage notes.

Our main backbones are Qwen3-VL-8B and Qwen3-VL-30B-A3B \cite{bai2025qwen3vltechnicalreport}. For table
instances, \name{} uses three calls: claim decomposition, table
grounding, and entailment resolution. Structured table JSON and
source-table text are provided when available as reading aids. For figure
instances, \name{} uses a reader--auditor--resolver workflow over the
current figure image and derived image views. Unless otherwise noted, we
use temperature 0.25 for tables, temperature 0.6 for figures,
top-\(p=0.95\), top-\(k=20\), and seed 1234. Outputs are normalized to
\textsc{Supported} or \textsc{Refuted}. We report macro-F1 over
individual instances and pair accuracy over paired groups.

\subsection{Results and Analysis}

\paragraph{Aggregate results.}
\autoref{tab:main-results} reports results for table evidence, figure
evidence, and the full development set. Across both backbones, \name{}
improves over direct multimodal prompting. With Qwen3-VL-30B-A3B,
\name{} achieves 79.2 macro-F1 and 63.1 pair accuracy overall, improving
over the vanilla baseline by 3.2 and 8.3 points, respectively. With
Qwen3-VL-8B, \name{} improves macro-F1 by 4.1 points and pair accuracy
by 8.5 points. The larger gain in pair accuracy suggests that staged
grounding improves evidence sensitivity, since paired examples require
the model to distinguish supporting from refuting evidence for the same
claim.

Performance differs by modality. Table verification is stronger overall:
with Qwen3-VL-30B-A3B, \name{} reaches 81.7 macro-F1 and 67.4 pair
accuracy on table evidence, reflecting the benefit of structured table
representations and row--column--cell grounding. Figure verification
remains harder but benefits substantially from the agentic workflow:
figure pair accuracy with the 30B backbone rises from 38.8 to 54.3,
indicating that explicit checks over panels, legends, categories, and
trends help reduce brittle direct judgments.

\paragraph{Efficiency.}
The two backbones show different generation costs under the same agentic
procedure. In our raw stage logs, Qwen3-VL-8B generates about 52.9K
tokens per table instance and 38.1K per figure instance, whereas
Qwen3-VL-30B-A3B generates about 26.5K and 23.2K tokens, respectively.
Since the prompt-side evidence packs are comparable, the difference
mainly comes from intermediate reasoning. This suggests that the larger
backbone executes the staged decomposition more compactly and uses the evidence more efficiently, especially for figure verification.

\paragraph{Operation-level results.}
\autoref{fig:operation-level-comparison} breaks down pair accuracy by
evidence modification type. For tables, \name{} improves most on changed
cell values: pair accuracy increases from 53.0 to 64.5 with Qwen3-VL-8B
and from 62.8 to 69.9 with Qwen3-VL-30B-A3B. Gains are smaller for
swapped rows or columns, suggesting that altered
numerical entries are well handled by row--column--cell grounding, but
row/column alignment remains difficult.

For figures, the largest gains occur when verification requires explicit
visual binding. With Qwen3-VL-30B-A3B, graph-flip pair accuracy rises
from 35.0 to 70.0, legend-swap accuracy rises from 39.7 to 54.0, and
graph-swap examples improve by 10.0 points for both backbones. However,
category swaps remain challenging: accuracy is unchanged for 8B at 34.8
and improves only from 39.1 to 43.5 for 30B. These results show that
\name{} improves binding to axes, legends, panels, and graph identity,
while fine-grained category matching and directional interpretation
remain brittle.

\section{Related Works}

\paragraph{Entailment Verification}
Entailment verification in the LLM era has shifted from static sentence-pair classification to retrieval-augmented, rationale-aware, and evidence-grounded reasoning over complex claims~\cite{thorne2018feverlargescaledatasetfact,wadden2022scifactopenopendomainscientificclaim}. Recent work studies how LLMs perform structured logical inference, proof construction, and entailment verification under diverse scenarios~\cite{han2024folionaturallanguagereasoning,zheng2025logidynamicsunravelingdynamicsinductive,zheng2025enhancingtransformersgeneralizablefirstorder,zheng2025cursecotlimitationschainofthought}. Parallel efforts extend verification beyond text by requiring models to reason over visual evidence, tables, figures, and multimodal contexts~\cite{deng2024texttupletableinformationintegrationtexttotable,mo2025dixitworldevaluatingmultimodalabductive,liang2025llmhanabievaluatingmultiagentgameplays}. 

\paragraph{LLM Agents in Scientific Discovery}
LLM-based scientific agents are increasingly involved in scientific discovery throughout the research lifecycle~\cite{zheng2025automationautonomysurveylarge,luo2025llm4srsurveylargelanguage}. Recent studies further push scientific discovery agents toward deeper autonomy, for tasks ranging from scientific deep research to scientific law discovery~\cite{zheng2026newtonbenchbenchmarkinggeneralizablescientific,shojaee2025llmsrscientificequationdiscovery,fang2026cognitivekernelproframeworkdeep,zheng2026sciresearcherscalingdeepresearch,xu2025probingscientificgeneralintelligence}. With the development of multi-modality in agentic systems, long-context vision-language modeling and persistent multimodal memory have become important for analyzing scientific documents and maintaining grounded evidence across extended trajectories~\cite{wang2025mmlongbenchbenchmarkinglongcontextvisionlanguage,ren2026memlensbenchmarkingmultimodallongterm,sun2026scienceboardevaluatingmultimodalautonomous}.

\section{Conclusion}
We presented \name{}, an agentic framework for scientific claim
verification with atomic claim decomposition and modality-specific
evidence grounding. On the SciClaimEval development set, \name{} improves
over vanilla multimodal prompting, achieving 79.2 macro-F1 and 63.1 pair
accuracy with Qwen3-VL-30B-A3B. The strongest gains occur in pair
accuracy and visually grounded figure cases, suggesting improved evidence
sensitivity, while remaining errors on row/column swaps, fine-grained
category matching, and directional visual reasoning highlight directions
for future work.

\section*{Acknowledgments}
The authors of this paper were supported by the ITSP Platform Research Project (ITS/189/23FP) from ITC of Hong Kong, SAR, China, and the AoE (AoE/E-601/24-N), the RIF (R6021-20) and the GRF (16205322) from RGC of Hong Kong, SAR, China. We also thank the support from NVIDIA AI Technology Center (NVAITC).

\bibliographystyle{ACM-Reference-Format}
\bibliography{sample-sigconf-xelatex}

\newpage
\appendix

\section{Prompt Details}

The full prompts used by \name{} are longer and include evidence-specific
content such as captions, selected table rows, source snippets, image
descriptions, and previous same-item stage notes. We show compact versions of
the reusable stage instructions below. All prompts are single-evidence prompts:
the model is not allowed to use pair identifiers, modification operations, gold
labels, paired evidence, previous predictions, or external knowledge.

\subsection{Shared Prompt Contract}

\begin{promptbox}{Evidence-Conditioned Verification Contract}
\textbf{Input.} One scientific claim, one current evidence item, caption,
and allowed auxiliary context.

\textbf{Evidence priority.} The current table or figure image is the decisive
evidence. Structured table JSON, source table text, caption, context, and
paper-focus text may help identify entities, units, panels, or terminology, but
they cannot override what is shown in the current evidence.

\textbf{Decision rule.} Predict \texttt{Supported} only when every central
empirical atom in the claim is entailed by the current evidence. Predict
\texttt{Refuted} if any central atom is contradicted, missing, unclear,
misaligned, or supported only by auxiliary text.
\end{promptbox}

\subsection{Table Agent Prompts}

\begin{promptbox}{Table Round 1: Claim Decomposer}
You are the claim-decomposition stage for an independent table verification
agent. Do not decide the final label.

\textbf{Task.} Split the claim into \emph{central empirical atoms} and
\emph{background atoms}. Central atoms must be table-checkable requirements
about entities, rows, columns, values, metrics, ranks, directions, counts,
arithmetic, significance, scope, or formatting cues. Background atoms are
mechanisms, motivations, interpretations, or definitions that the table need not
prove unless the evidence directly contradicts them.

\textbf{Strictness.} Keep exact numbers, named metrics, row/column bindings,
rank terms, and hard scope terms strict. Mark approximate or qualitative
wording separately so later stages can judge direction and reasonable
magnitude.

\textbf{Return JSON fields.} \texttt{central\_atoms},
\texttt{background\_atoms}, and \texttt{decomposition\_notes}. Each central
atom includes \texttt{atom}, \texttt{atom\_type}, \texttt{strictness}, and
\texttt{needed\_evidence}.
\end{promptbox}

\begin{promptbox}{Table Round 2: Evidence Grounder}
You are the evidence-grounding stage for one current table item. Use the Round 1
atoms as a checklist.

\textbf{Task.} For each central atom, align it to concrete table evidence:
row name, column name, cell value, unit, formatting cue, full-scope inspection,
or arithmetic expression. Structured JSON and source text only help read the
current table; caption/context cannot prove cell values.

\textbf{Checks.} For exact values and row--column bindings, cite the specific
cell. For \texttt{all}, \texttt{only}, \texttt{best}, \texttt{highest}, or
\texttt{consistently}, inspect the relevant full scope. For averages, ratios,
differences, or percent-point claims, write the calculation.

\textbf{Return JSON fields.} \texttt{atom\_checks},
\texttt{contradictions}, \texttt{missing\_or\_unclear}, and
\texttt{grounding\_notes}. Each atom check includes \texttt{atom},
\texttt{status}, \texttt{evidence}, and optional \texttt{calculation}.
\end{promptbox}

\begin{promptbox}{Table Round 3: Entailment Resolver}
You are the final table entailment resolver. Use the evidence pack, claim
decomposition, and grounding notes for this same evidence item only.

\textbf{Resolve.} \texttt{Supported} if all central empirical atoms are entailed
by the current table and no central contradiction remains. \texttt{Refuted} if
any central atom is contradicted, missing, unclear, misaligned, or supported
only by caption/context/paper text. Background atoms do not cause
\texttt{Refuted} unless the table directly contradicts them.

\textbf{Return JSON fields.} \texttt{verdict}, \texttt{brief\_reason},
\texttt{decisive\_atoms}, and optional \texttt{resolver\_checks}.
\end{promptbox}

\subsection{Figure Agent Prompts}

\begin{promptbox}{Figure Round 1: Visual Reader}
You are the figure reader for an independent scientific claim verification
agent. Inspect one claim and one current figure only.

\textbf{Task.} Convert the claim into atomic visual checks. Read the relevant
panel, title, axes, axis direction, legends, colors, line styles, markers,
categories, values, trends, peaks, and rankings from the current image. Return a
provisional label, but prioritize visual facts, contradictions, and uncertainty.

\textbf{Return JSON fields.} \texttt{claim\_atoms}, \texttt{visual\_facts},
\texttt{contradictions}, \texttt{uncertain}, \texttt{provisional\_label}, and
\texttt{brief\_reason}.
\end{promptbox}

\begin{promptbox}{Figure Round 2: Evidence-Binding Auditor}
You are the evidence-binding auditor for one current figure item.

\textbf{Task.} Before judging the claim, bind each compared entity through the
current image: panel $\rightarrow$ axis/metric $\rightarrow$
legend/color/marker/line/label $\rightarrow$ observed value, trend, rank, peak,
or direction. Do not infer colors, categories, rankings, or values from the
claim or caption.

\textbf{Qualifier policy.} Hard logical qualifiers such as \texttt{all},
\texttt{every}, \texttt{only}, \texttt{highest}, \texttt{lowest},
\texttt{same}, and \texttt{during all} are strict. Statistical qualifiers
require visual markers or reported values when the claim explicitly depends on
significance or confidence intervals. Soft qualifiers such as
\texttt{slightly}, \texttt{similar}, \texttt{better}, and \texttt{more
pronounced} are judged by visible trend or relative gap.

\textbf{Return JSON fields.} \texttt{label}, \texttt{brief\_reason},
\texttt{atom\_checks}, \texttt{binding\_checks}, \texttt{qualifier\_checks},
\texttt{contradictions}, and \texttt{uncertain}.
\end{promptbox}

\begin{promptbox}{Figure Round 3: Visual Entailment Resolver}
You are the final resolver for one current figure item. Use only this claim,
this evidence, and previous same-evidence stage notes.

\textbf{Resolve.} Identify the figure-checkable empirical core: visual
comparison, trend, rank, peak, direction, category relation, numeric relation,
and truth-changing qualifier. Predict \texttt{Supported} only if current-image
bindings are correct, the core visual assertion is supported, hard qualifiers
hold, required statistical evidence is visible when needed, and no central
contradiction remains. Predict \texttt{Refuted} for binding failures,
panel/metric/category mismatches, wrong direction/rank/peak, missing or
unreadable central evidence, or failed hard/required qualifiers.

\textbf{Return JSON fields.} \texttt{label} and \texttt{brief\_reason}.
\end{promptbox}

\end{document}